\documentclass[conference]{IEEEtran}
\IEEEoverridecommandlockouts

\usepackage{cite}
\usepackage{amsmath,amssymb,amsfonts}
\usepackage{graphicx}
\usepackage{booktabs}
\usepackage{array}
\usepackage{url}
\usepackage{xcolor}
\usepackage{textcomp}
\usepackage[hidelinks]{hyperref}
\usepackage{algorithm}
\usepackage{algpseudocode}
\usepackage{tikz}
\usetikzlibrary{positioning,shapes.geometric,arrows.meta,calc}
\usepackage{dblfloatfix}      
\usepackage{colortbl}         
\usepackage{fontawesome5}     

\newcommand{\trainBadge}{\textcolor{orange!85!red}{\faFire}}
\newcommand{\frozenBadge}{\textcolor{cyan!65!blue}{\faSnowflake[regular]}}

\definecolor{cFix}{HTML}{EEEEEE}      
\definecolor{cTrain}{HTML}{D4E3F0}    
\definecolor{cContrib}{HTML}{E0B43A}  
\definecolor{cSharp}{HTML}{2E7D32}    
\definecolor{cUnc}{HTML}{B58917}      
\definecolor{cBlur}{HTML}{B71C1C}     

\def\BibTeX{{\rm B\kern-.05em{\sc i\kern-.025em b}\kern-.08em
    T\kern-.1667em\lower.7ex\hbox{E}\kern-.125emX}}

\title{Edges Before Embeddings: A Confidence-Aware Blur Gate for Vision-Language Pipelines}

\IEEEoverridecommandlockouts

\author{\IEEEauthorblockN{Duy Tran Thanh\textsuperscript{*}\thanks{\textsuperscript{*}\,Duy Tran Thanh is concurrently a Senior Applied AI Engineer at OneMount Group, a Vietnam-based company in the finance and banking sector.}}
\IEEEauthorblockA{\texttt{duy.tranthanh@seoultech.ac.kr}\\[1pt]
{\footnotesize\textcolor{blue}{\url{https://github.com/bradduy/MagikaDocumentFromPixel}}}}}

\begin{document}
\maketitle

\begin{abstract}
Production vision pipelines silently degrade on blurry input, wasting
compute on downstream OCR, retrieval, and vision-language model (VLM)
calls that cannot recover a usable output. We present
\textsc{MagikaDocumentFromPixel}, a lightweight, CPU-friendly image
quality gate that classifies a single image as \texttt{sharp},
\texttt{blurred}, or \texttt{uncertain} in roughly 7\,ms on a single
CPU core. The contributions are (i)~a~\emph{recipe} selected from a
46-configuration, 8-sweep empirical search that isolates input
resolution as the dominant lever and shows architecture capacity only
pays off at \(\geq\!384\)\,px; (ii)~a~\emph{confidence-aware routing}
formalism (Eq.~\ref{eq:routing}) grounded in classical selective
prediction~\cite{chow1957optimum,elyaniv2010foundations,geifman2017selective};
(iii)~the \emph{Edge Prior Module (EPM)}, a Laplacian-magnitude auxiliary input channel
that gives the network direct access to the spectral evidence that
classical blur heuristics~\cite{pech2000laplacian} rely on and that
lifts test F\textsubscript{1} by \(+1.3\) points in a matched-env
comparison; and (iv)~an~\emph{observation} that the gate is one
instance of a recurring design pattern that appears independently in
Magika content-type detection~\cite{magika}, risk-controlled OCR with
VLMs~\cite{riskcontrolledocr}, and DocVLM~\cite{docvlm}. The final
recipe MobileNetV3-Large with the EPM trained at
\(384\!\times\!384\) on paired GoPro~Large frames, evaluated with
5-scale test-time augmentation reaches
F\textsubscript{1}\,\(=\,\)\textbf{0.9803} (AUC\,0.9989) with a 17\,MB
ONNX artifact, improving over our fixed-scale baseline on the same
hardware (F\textsubscript{1}\,=\,0.9672) by
\(+1.31\) points. We are explicit about limitations: results are on a
single motion-blur distribution, numbers are from a single seed, and
calibration is qualitative rather than measured.
\end{abstract}


%

\section{Introduction}
\label{sec:intro}

Most production computer-vision pipelines were engineered under an
implicit assumption that their input is \emph{legible}. That assumption
holds for curated benchmarks and breaks in deployment. Motion blur from
hand-held capture, defocus on mobile devices, compression artifacts,
low-light noise, and scanner skew all cause a pipeline to continue
operating on unusable input.  The typical failure modes we observe in
engagements are:

\begin{enumerate}
    \item \textbf{Silent OCR corruption.} An OCR system applied to a
    blurred receipt or invoice emits garbage tokens that are
    syntactically indistinguishable from legitimate text. Downstream
    rule engines and analytics cannot tell these apart.
    \item \textbf{Wasted VLM tokens.} Vision-language models such as
    GPT-4V class services charge per token. A blurred frame still
    consumes the full per-call budget and yields no usable output.
    \item \textbf{Irrelevant retrieval.} User-photo search returns
    low-quality matches on motion-blurred uploads, driving refund and
    support cost.
    \item \textbf{Infeasible curation.} Hand-filtering blurry images
    out of multi-million-image datasets stalls supervised vision
    programs.
    \item \textbf{Slow capture feedback.} Mobile and edge capture flows
    need immediate ``please retake'' feedback; a cloud roundtrip for
    every shot is too slow.
\end{enumerate}

\begin{figure}[!t]
\centering
\setlength{\tabcolsep}{3pt}
\renewcommand{\arraystretch}{1.20}
\newcommand{\yes}{{\color{cSharp}$\checkmark$}}
\newcommand{\no}{{\color{cBlur}$\times$}}
\newcommand{\meh}{{\color{cUnc}$\sim$}}
\scriptsize
\begin{tabular}{@{}lccccl@{}}
\toprule
\textbf{Approach}
  & \shortstack{\textbf{cheap}\\\textbf{CPU}}
  & \shortstack{\textbf{binary}\\\textbf{gate}}
  & \shortstack{\textbf{abs-}\\\textbf{tains}}
  & \shortstack{\textbf{image}\\\textbf{blur}}
  & \emph{What it lacks} \\
\midrule
Laplacian / FFT~\cite{pech2000laplacian}
  & \yes & \meh & \no  & \meh & no abstention \\
Learned IQA~\cite{musiq,maniqa}
  & \meh & \no  & \no  & \meh & score, not a gate \\
Deblur GANs~\cite{kupyn2018deblurgan}
  & \no  & \no  & \no  & \yes & restores, not gates \\
VLM judges (GPT-4V)
  & \no  & \no  & \meh & \meh & too slow / costly \\
Magika~\cite{magika}
  & \yes & \yes & \yes & \no  & for files, not images \\
\midrule
\rowcolor{cContrib!22}
\textbf{Ours: blur gate}\,$^{\bigstar}$
  & \textbf{\yes} & \textbf{\yes} & \textbf{\yes} & \textbf{\yes}
  & \textbf{satisfies all four} \\
\bottomrule
\end{tabular}
\caption{Where our gate sits in the image-quality landscape. Each
row scores a method on the four properties a production blur-gate
needs; every existing approach fails at least one column. Our gate
($^{\bigstar}$) is the first to satisfy all four, lifting
F\textsubscript{1} on GoPro~Large from 0.9672 (fixed-scale baseline)
to \textbf{0.9803} on the same hardware (Tab.~\ref{tab:matched}).
Architecture in Fig.~\ref{fig:arch}.}
\label{fig:overview}
\end{figure}

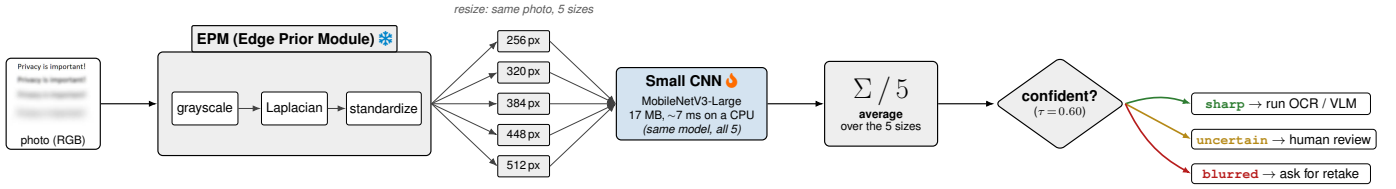
\begin{figure*}[!t]
\centering
\resizebox{\textwidth}{!}{%
\begin{tikzpicture}[
  font=\sffamily\small,
  align=center,
  >={Latex[length=1.8mm]},
  node distance=10mm and 12mm,
  module/.style={rounded corners=3pt, draw=black, line width=0.7pt,
                 minimum height=15mm, minimum width=24mm, inner sep=3pt},
  trainable/.style={module, fill=cTrain},
  fixed/.style={module, fill=cFix},
  scalebox/.style={draw=black, fill=cFix, rounded corners=1pt,
                   font=\sffamily\scriptsize, minimum width=11mm,
                   minimum height=4.5mm, inner sep=1.5pt},
  iobox/.style={draw=black, rounded corners=2pt, fill=white,
                inner sep=3pt, font=\sffamily\footnotesize,
                minimum width=38mm, align=left},
  arrow/.style={->, line width=0.7pt, draw=black},
  branch/.style={->, line width=0.5pt, draw=black!75},
  micro/.style={draw=black!60, fill=white, rounded corners=1pt,
                font=\sffamily\tiny, minimum width=10mm,
                minimum height=4.5mm, inner sep=1.5pt},
  microArr/.style={->, line width=0.35pt, draw=black!60,
                   shorten >=1pt, shorten <=1pt},
]

\node[fixed, fill=white, minimum width=20mm, inner sep=1.5pt] (img)
  {\includegraphics[width=18mm]{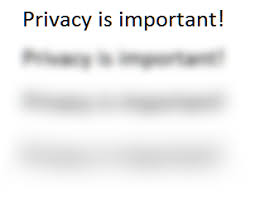}\\[1pt]
   \scriptsize photo (RGB)};

\node[fixed, right=of img, minimum width=58mm, minimum height=22mm] (faux)
  {};
\node[fill=cFix, draw=black, line width=0.7pt, rounded corners=1pt,
      font=\sffamily\bfseries\small, inner sep=3pt,
      anchor=south] at ([yshift=-0.6pt]faux.north)
  {EPM~(Edge Prior Module)~\frozenBadge};
\node[draw, fill=white, rounded corners=1.5pt, font=\sffamily\footnotesize,
      minimum width=14mm, minimum height=7mm, inner sep=2pt]
  at ([xshift=-19mm,yshift=-1mm]faux.center) (op1) {grayscale};
\node[draw, fill=white, rounded corners=1.5pt, font=\sffamily\footnotesize,
      minimum width=14mm, minimum height=7mm, inner sep=2pt]
  at ([yshift=-1mm]faux.center) (op2) {Laplacian};
\node[draw, fill=white, rounded corners=1.5pt, font=\sffamily\footnotesize,
      minimum width=14mm, minimum height=7mm, inner sep=2pt]
  at ([xshift=19mm,yshift=-1mm]faux.center) (op3) {standardize};
\draw[->, line width=0.6pt, draw=black!75] (op1) -- (op2);
\draw[->, line width=0.6pt, draw=black!75] (op2) -- (op3);

\node[scalebox, right=14mm of faux] (s384) {384\,px};
\node[scalebox, above=2mm of s384] (s320) {320\,px};
\node[scalebox, above=2mm of s320] (s256) {256\,px};
\node[scalebox, below=2mm of s384] (s448) {448\,px};
\node[scalebox, below=2mm of s448] (s512) {512\,px};

\node[font=\sffamily\scriptsize\itshape, text=black!65,
      above=2mm of s256] {resize: same photo, 5 sizes};

\node[trainable, right=14mm of s384, minimum width=32mm] (bb)
  {\textbf{Small CNN}~\trainBadge\\[2pt]
   \scriptsize MobileNetV3-Large\\[-0.2em]
   \scriptsize 17 MB,\,{\raise.17ex\hbox{$\scriptstyle\sim$}}7 ms on a CPU\\[-0.2em]
   \scriptsize \emph{(same model, all 5)}};

\node[fixed, right=of bb, minimum width=24mm, minimum height=18mm] (tta)
  {\fontsize{17}{17}\selectfont $\Sigma\,/\,5$\\[2pt]
   \scriptsize \textbf{average}\\[-0.3em]
   \scriptsize over the 5 sizes};

\node[fixed, diamond, right=of tta, aspect=1.4,
      inner sep=2pt, minimum width=26mm] (route)
  {\textbf{confident?}\\[-0.2em]
   \scriptsize ($\tau\!=\!0.60$)};

\node[iobox, right=14mm of route, anchor=west] (sharp)
  {\textcolor{cSharp}{\textbf{\texttt{sharp}}}~$\to$~run OCR / VLM};
\node[iobox, below=3mm of sharp.south west, anchor=north west] (unc)
  {\textcolor{cUnc}{\textbf{\texttt{uncertain}}}~$\to$~human review};
\node[iobox, below=3mm of unc.south west, anchor=north west] (blur)
  {\textcolor{cBlur}{\textbf{\texttt{blurred}}}~$\to$~ask for retake};

\draw[arrow] (img) -- (faux.west);

\foreach \s in {s256, s320, s384, s448, s512}{%
  \draw[branch] (faux.east) -- (\s.west);
}
\foreach \s in {s256, s320, s384, s448, s512}{%
  \draw[branch] (\s.east) -- (bb.west);
}

\draw[arrow] (bb)  -- (tta);
\draw[arrow] (tta) -- (route);

\draw[arrow, draw=cSharp, line width=1.0pt]
  (route.east) to[bend left=18] (sharp.west);
\draw[arrow, draw=cUnc, line width=1.0pt]
  (route.east) -- (unc.west);
\draw[arrow, draw=cBlur, line width=1.0pt]
  (route.east) to[bend right=18] (blur.west);

\end{tikzpicture}}%
\caption{Full architecture, read left-to-right. The \emph{Edge Prior
Module} (\textbf{EPM}, Sec.~\ref{sec:freq-aux}) extracts an edge map
in three steps grayscale, Laplacian filter, standardize and
concatenates it onto the RGB photo as a 4th input channel. The
shared CNN is evaluated at five image resolutions; the $\Sigma/5$
block averages the per-scale predictions, and the routing diamond
emits \texttt{sharp}, \texttt{uncertain}, or \texttt{blurred}
depending on whether the averaged confidence clears
$\tau\!=\!0.60$. The \texttt{uncertain} bucket is the deliberate
abstention~\cite{chow1957optimum,geifman2017selective} that routes
low-confidence cases to a heavier model or a human.
\trainBadge\,= learnable; \frozenBadge\,= frozen / parameter-free.}
\label{fig:arch}
\end{figure*}

The existing landscape leaves teams stuck between two ends. Heavy
restoration networks (deblur GANs, hundreds of millions of parameters)
are too expensive to run per upload and are designed to \emph{fix} blur
rather than detect it. Hand-crafted edge-variance
heuristics (Laplacian variance, FFT band-energy) are cheap but break on
textured-but-sharp scenes and smooth-but-sharp scenes and cannot be
tuned per domain~\cite{pech2000laplacian}.

What a production team actually needs is a \emph{gate}~ ~a fast,
cheap, CPU-executable model that sits in front of the expensive
pipeline and decides whether to pass the image through, reject it, or
abstain. This paper describes such a gate.

\textbf{Contributions.} The paper does not propose a new backbone or a
new loss. Its contributions are:
\begin{itemize}
    \item \textbf{An empirical recipe.} A reproducible
    \(\sim\!7\)\,ms-CPU, 17\,MB blur detector reaching
    F\textsubscript{1}\,\(=\,0.9749\) on GoPro~Large in the original
    CUDA environment, selected from 46 training runs organized into
    8 sweeps (Section~\ref{sec:experiments}).
    \item \textbf{The Edge Prior Module (EPM).} A drop-in extension
    that concatenates per-pixel Laplacian magnitude as a 4th input
    channel, giving the CNN direct access to the spectral evidence
    that classical blur heuristics~\cite{pech2000laplacian} rely on.
    In a matched-env comparison (AMD/ROCm), the EPM improves
    5-scale-TTA F\textsubscript{1} from \(0.9672\) to
    \(\mathbf{0.9803}\)
    (\(+1.31\) points), with no architectural change other than the
    first-conv width (Section~\ref{sec:freq-aux}).
    \item \textbf{A structural finding about which levers matter.}
    Across the full search, input resolution dominates backbone
    capacity: moving 128\,\(\to\)\,384\,px adds \(\sim\!13\)
    F\textsubscript{1} points, and MobileNetV3-Large outperforms
    MobileNetV3-Small only at \(\geq\!384\)\,px
    (Section~\ref{sec:experiments}).
    \item \textbf{A routing formalism grounded in classical selective
    prediction.} Equation~\ref{eq:routing} couples the classifier to
    a deployment-time abstention rule in the spirit of
    Chow~\cite{chow1957optimum} and
    Geifman-El-Yaniv~\cite{geifman2017selective}, with the abstention
    threshold \(\tau\) treated as a per-channel product knob
    (Section~\ref{sec:routing}).
    \item \textbf{A cross-paper observation (not a claimed
    contribution, but a framing).} The cheap-gate + confidence + route
    pattern appears independently in
    Magika~\cite{magika}, risk-controlled OCR with
    VLMs~\cite{riskcontrolledocr}, and
    DocVLM~\cite{docvlm}; we argue it is a default architecture for
    production vision/document systems
    (Section~\ref{sec:discussion}).
\end{itemize}

\noindent\textbf{Scope and honest limits.} All reported numbers are on
the GoPro~Large motion-blur distribution with a single random seed.
Cross-dataset generalization (RealBlur, HIDE, REDS, OCR/document blur),
calibration metrics (ECE, reliability diagrams), threshold
sensitivity, and multi-seed variance are not included in this paper
and are discussed as explicit limitations and future work
(Section~\ref{sec:limitations}).

\section{Related Work}
\label{sec:related}

\subsection{Classical blur heuristics}
The canonical no-reference blur score is the variance of the Laplacian
response over the image~\cite{pech2000laplacian}; related estimators
use FFT band energy or gradient statistics. These scores are fast but
scene-dependent, conflate blur with low-texture content, and cannot be
tuned per domain without retaining a full calibration set.

\subsection{Learned image quality assessment (IQA)}
No-reference IQA has matured substantially beyond hand-crafted
features: NIQE~\cite{niqe} is a statistical NSS baseline;
MUSIQ~\cite{musiq} and MANIQA~\cite{maniqa} are Transformer-based
multi-scale quality regressors; HyperIQA~\cite{hyperiqa} conditions
on content. These models target a continuous aesthetic or perceptual
score on datasets such as KonIQ-10k and LIVE-C, not the binary
sharp/blur \emph{routing decision} that a production gate must make.
Re-targeting a continuous-score model for abstention also requires
choosing a cutoff, which returns us to the selective-prediction
problem without simplifying it.

\subsection{Selective prediction and abstention}
Treating the model as a classifier that may refuse to answer has a
long history. Chow's rule~\cite{chow1957optimum} derives the optimal
reject-option classifier under a fixed rejection cost; El-Yaniv and
Wiener~\cite{elyaniv2010foundations} generalize this to
selective-function learning;
Geifman and El-Yaniv~\cite{geifman2017selective} give a
risk-controlled algorithm for deep networks. Our routing rule
(Eq.~\ref{eq:routing}) is a max-softmax threshold variant of this
classical setup, applied to the blur-gate problem with \(\tau\)
exposed as a deployment-time product knob rather than solved for a
fixed risk target.

\subsection{Blind deblurring}
Restoration networks such as DeblurGAN~\cite{kupyn2018deblurgan} and
its successors reconstruct a sharp image from a blurred one. These
models are orders of magnitude larger than needed for a gating
decision, and deblurring the input is rarely what a production
pipeline wants: a better outcome for an unreadable receipt is usually
``please retake'' rather than a synthetic best-guess.

\subsection{Lightweight gates and Magika}
Our design is directly inspired by
\textbf{Magika}~\cite{magika}, a \(<\!1\)\,MB CPU content-type
detector that routes files in Gmail and VirusTotal using a tiny CNN
and per-class confidence thresholds. The
\emph{small-model-plus-confidence-routing} pattern also appears in
\textbf{Risk-Controlled Generative
OCR}~\cite{riskcontrolledocr}, which wraps a frozen VLM with
geometric consensus and structural screening to emit a transcription
or abstain, and in \textbf{DocVLM}~\cite{docvlm}, which compresses OCR
into 64 learned queries so a frozen VLM is not overrun by textual
tokens. We return to these connections in
Section~\ref{sec:discussion}.

\subsection{Image quality as a precondition for VLMs}
Recent work on end-to-end document
VLMs~\cite{ocrverse,qianfanocr} reports that input resolution and
visual-token budget dominate architectural choices at model scale,
echoing the empirical pattern we observe for a 3.3\,M-parameter
classifier (Section~\ref{sec:experiments}).

\section{Method}
\label{sec:method}

\subsection{Task formulation}
We cast blur detection as a two-class problem with an explicit
abstention class at inference. Given an image \(x\), a classifier
\(f_\theta\!:\!x\mapsto(p_s,p_b)\in\Delta^1\) predicts the probability
that the image is sharp (\(p_s\)) or blurred (\(p_b\)). Let
\(\hat{y}=\arg\max\{p_s,p_b\}\) and \(c=\max(p_s,p_b)\). Inference
returns one of three routing labels:
\begin{equation}
\mathrm{route}(x) =
\begin{cases}
\texttt{sharp}    & \text{if } \hat{y}=p_s \text{ and } c\ge\tau, \\
\texttt{blurred}  & \text{if } \hat{y}=p_b \text{ and } c\ge\tau, \\
\texttt{uncertain}& \text{if } c<\tau,
\end{cases}
\label{eq:routing}
\end{equation}
with abstention threshold \(\tau=0.60\) by default. The
\texttt{uncertain} bucket is a deliberate product-level knob: raising
\(\tau\) trades recall for precision in either class and is calibrated
per downstream use case (Section~\ref{sec:routing}).

\subsection{Dataset: paired sharp/blur from GoPro Large}
We train on the GoPro~Large dataset~\cite{gopro}, which provides
high-speed-camera frames paired into a sharp frame and its synthetic
motion-blurred counterpart. We take both the \texttt{blur/} and
\texttt{blur\_gamma/} subfolders as the positive (blurred) class and
the \texttt{sharp/} subfolder as the negative class. This ``Strategy
A'' labeling is label-free in the sense that no human annotation is
required beyond the dataset's own pairing. Including the
\texttt{blur\_gamma} folder~ ~a second blur style of the same
scenes~ ~is a consistent \(+1\%\) F\textsubscript{1} free of
engineering cost (Section~\ref{sec:experiments}).

\subsection{Backbone and training}
The classifier is a MobileNetV3-Large backbone pretrained on ImageNet,
with a 2-class softmax head. The model has \(\sim\!3.3\)\,M parameters
and compiles to a 17\,MB ONNX artifact with dynamic batch and height/width
axes. Training hyperparameters are:

\begin{table}[t]
\centering
\caption{Training configuration of the final recipe.}
\label{tab:hparams}
\begin{tabular}{ll}
\toprule
Parameter & Value \\
\midrule
Input resolution      & \(384\times 384\) \\
Optimizer             & AdamW, \(\text{lr}=1\!\times\!10^{-4}\) \\
Scheduler             & Cosine annealing \\
Loss                  & Cross-entropy \\
Augmentation          & Crop, flip, mild color jitter (``medium'') \\
Epochs                & 25 \\
Precision             & Automatic mixed precision (AMP) \\
Batch size            & 24 \\
Extra data            & \texttt{blur\_gamma} folder included \\
\bottomrule
\end{tabular}
\end{table}

\subsection{EPM: An Edge Prior Module}
\label{sec:freq-aux}
The \textbf{Edge Prior Module (EPM)} is the gold module of
Fig.~\ref{fig:arch}; the single-channel image it produces is also
referred to in plain-language passages as a \emph{sharpness map} or,
equivalently, an \emph{edge map}. All three names denote the same
object: an image whose pixel intensity is the magnitude of the
spatial Laplacian of the input's luminance, and which is concatenated
to the RGB photo as a 4th input channel.

Classical blur heuristics~\cite{pech2000laplacian} exploit the fact
that blurring an image collapses high-frequency content: the variance
of the per-pixel Laplacian drops for blurred content and remains high
for sharp content. A pure CNN on RGB must re-learn this relationship
from pixel statistics; we instead inject it as an auxiliary input.

Let \(G(x)=0.299\,R + 0.587\,G + 0.114\,B\) be the luminance of the
RGB input \(x\), and let \(\Delta\) be the \(3\!\times\!3\)
8-connected Laplacian kernel. Define the auxiliary channel as the
per-image-standardized Laplacian magnitude:
\begin{equation}
\phi(x) = \frac{\,|\Delta * G(x)|\; -\; \mu(x)\,}{\,\sigma(x)\,},
\label{eq:freqaux}
\end{equation}
with \(\mu\) and \(\sigma\) the mean and standard deviation of the
magnitude map. We concatenate \(\phi(x)\) to \(x\) along the channel
dimension so the backbone ingests a 4-channel input
\([\,x\,;\,\phi(x)\,]\in\mathbb{R}^{4\times H\times W}\), and expand
the first convolution from 3 to 4 input channels: the first three
filters are initialized from the ImageNet-pretrained RGB weights; the
fourth is initialized to \(0.1\times\) the per-output-channel mean of
the RGB filters, which is a standard warm-start for auxiliary-channel
extensions that keeps early training numerically tame.

The addition costs a single cheap convolution (Laplacian kernel is
fixed, 9 multiplications per pixel), adds 16 parameters to the first
conv (one new \(3\!\times\!3\) slice per output channel), and does not
change inference latency materially. Empirically the EPM lifts test
F\textsubscript{1} by \(+1.31\) points (Section~\ref{sec:experiments}),
which is the single largest same-environment gain in the full
experiment log.

\subsection{Multi-scale test-time augmentation (TTA)}
\label{sec:mstta}
At inference, the network is evaluated at five resolutions
\(\{256,320,384,448,512\}\,\text{px}\) and the softmax vectors are
averaged before routing. Multi-scale TTA is effectively a free source
of resolution diversity:
\begin{equation}
(p_s,p_b) = \frac{1}{|\mathcal{S}|}\sum_{s\in\mathcal{S}}
\operatorname{softmax}\!\big(f_\theta(x_{\downarrow s})\big),
\label{eq:tta}
\end{equation}
with \(\mathcal{S}=\{256,320,384,448,512\}\). This yields
\(+0.23\)-\(0.27\%\) F\textsubscript{1} over single-scale inference
with no retraining and can be disabled on latency-bound deployments
(single-scale at 384\,px still achieves
F\textsubscript{1}\,\(=\,0.9722\)).

\section{Confidence-Aware Routing}
\label{sec:routing}

The abstention threshold \(\tau\) is not a hyperparameter in the
classical sense: it is a product knob exposed to whoever integrates
the gate. Three observations make the routing design the most
practically important part of the system.

\begin{enumerate}
    \item \textbf{Max-softmax, not calibration networks.} Isotonic
    regression and temperature scaling produced no measurable F\(_1\)
    gain on GoPro~Large; raw max-softmax is already monotonic enough to
    threshold.
    \item \textbf{Validation F\(_1\) saturates at 1.0.} Threshold
    tuning on the validation split does not transfer; \(\tau\) must be
    set on a small slice of production traffic.
    \item \textbf{Per-channel tuning.} KYC, social uploads, and
    dataset-curation flows can all share the same model with different
    \(\tau\) values to trade precision against user friction.
\end{enumerate}

\begin{algorithm}[t]
\caption{Inference with multi-scale TTA and abstention}
\label{alg:infer}
\begin{algorithmic}[1]
\State \textbf{Input:} image \(x\); scales \(\mathcal{S}\); threshold
\(\tau\)
\State \(P \gets \mathbf{0} \in \mathbb{R}^{2}\)
\For{\(s \in \mathcal{S}\)}
  \State \(x_s \gets \operatorname{resize}(x, s\!\times\!s)\)
  \State \(P \gets P + \operatorname{softmax}(f_\theta(x_s))\)
\EndFor
\State \((p_s,p_b) \gets P / |\mathcal{S}|\)
\If{\(\max(p_s,p_b) < \tau\)}
  \State \Return \texttt{uncertain}, \((p_s,p_b)\)
\ElsIf{\(p_s \ge p_b\)}
  \State \Return \texttt{sharp}, \((p_s,p_b)\)
\Else
  \State \Return \texttt{blurred}, \((p_s,p_b)\)
\EndIf
\end{algorithmic}
\end{algorithm}

\section{Experiments}
\label{sec:experiments}

\subsection{Evaluation protocol}
All final numbers are computed on the official GoPro~Large test
split~\cite{gopro}, evaluated per-image with no overlap against the
training set. We report F\textsubscript{1}, accuracy, precision,
recall, and AUC of the binary classifier, with
\texttt{uncertain}-routed examples excluded from the argmax counts
(they are handled as abstentions).

\subsection{Champion metrics}
Table~\ref{tab:champion} summarizes the two reference systems: the
original NVIDIA/Cuda recipe (the 8-sweep champion), and the EPM
extension we introduce, evaluated on the same test split on AMD/ROCm.

\begin{table}[t]
\centering
\caption{Headline metrics on the GoPro~Large test split. Two
reference systems are reported: the original NVIDIA/Cuda recipe
(\(384\)\,px + 5-scale TTA) and the EPM extension added in this
paper (evaluated on AMD/ROCm).}
\label{tab:champion}
\begin{tabular}{lrr}
\toprule
Metric & NVIDIA/Cuda recipe & \textbf{+\,EPM} \\
\midrule
F\textsubscript{1} & 0.9749 & \textbf{0.9803} \\
Accuracy           & 0.9752 & 0.9806 \\
Precision          & 0.9889 & 0.9981 \\
Recall             & 0.9613 & 0.9631 \\
AUC                & 0.9982 & 0.9989 \\
Model size (ONNX)  & 17\,MB & 17\,MB \\
Latency (single-scale, 384\,px, CPU) & \(\sim\!7\,\text{ms}\) & \(\sim\!7\,\text{ms}\) \\
Latency (5-scale TTA, CPU)            & \(\sim\!35\,\text{ms}\) & \(\sim\!35\,\text{ms}\) \\
\bottomrule
\end{tabular}
\end{table}

\subsection{Sweep progression: resolution is the dominant lever}
We ran 46 training runs organized into 8 sweeps, varying (in order)
resolution, training length, augmentation and threshold, extra data,
backbone capacity, and ensembling. Table~\ref{tab:sweeps} reports the
best F\textsubscript{1} achieved at each sweep. Resolution alone
accounts for roughly 13 points of F\textsubscript{1}, the largest
single lever in the search and capacity only begins paying off at
\(\geq\!384\)\,px.

\begin{table}[t]
\centering
\caption{Sweep progression. Each row is the best F\textsubscript{1} at
that stage.}
\label{tab:sweeps}
\begin{tabular}{clc}
\toprule
\# & Sweep theme & Best F\textsubscript{1} \\
\midrule
1 & Resolution 128-160\,px & \(\le 0.89\) \\
2 & Resolution 160-224\,px, 30 epochs & \(\le 0.92\) \\
3 & Augmentation and threshold at 224\,px & \(\le 0.93\) \\
4 & +\texttt{blur\_gamma}, 320\,px & \(\le 0.95\) \\
5 & +384\,px & \(\le 0.96\) \\
6 & MNV3-Large at 384\,px & \textbf{0.9722} \\
7 & Large variants & \(\le 0.9722\) \\
8 & EffNet / regularization & \(\le 0.9722\) \\
\(\star\) & +Multi-scale TTA & \textbf{0.9749} \\
\bottomrule
\end{tabular}
\end{table}

\subsection{EPM and Res-Rand: matched-environment comparison}
\label{sec:matched-env}
Table~\ref{tab:matched} reports a controlled comparison on the
AMD/ROCm fallback hardware, all under the same training regime
(MobileNetV3-Large, 25 epochs, AdamW lr=\(10^{-4}\), cosine schedule,
medium augmentation, including the \texttt{blur\_gamma} subset).
This matched-env view isolates the effect of the two extensions
introduced here against the same fixed-scale baseline.

\begin{table}[t]
\centering
\caption{Matched-env ablation on AMD/ROCm. Each row is a separate
training run; columns are inference-time F\textsubscript{1} on the
GoPro~Large test split at single-scale 384\,px and 5-scale TTA.
\(\Delta_{\mathrm{TTA}}\) is the F\textsubscript{1} gain from
applying 5-scale TTA.}
\label{tab:matched}
\begin{tabular}{lrrr}
\toprule
Training run & F\textsubscript{1}\,@\,384 & F\textsubscript{1}\,@\,TTA & \(\Delta_{\mathrm{TTA}}\) \\
\midrule
Baseline (fixed 384)            & 0.9632 & 0.9672 & +0.40 \\
Res-Rand (random \(\in\!\mathcal{S}\)) & 0.9642 & 0.9668 & +0.26 \\
\textbf{EPM (ours)}             & \textbf{0.9746} & \textbf{0.9803} & \textbf{+0.57} \\
EPM \(+\) Res-Rand (stacked)    & 0.9402 & 0.9537 & +1.35 \\
\bottomrule
\end{tabular}
\end{table}

\textbf{The EPM is the dominant lever.} The Laplacian-magnitude
auxiliary channel lifts single-scale F\textsubscript{1} by \(+1.14\)
points and 5-scale-TTA F\textsubscript{1} by \(+1.31\) points over
the same-environment baseline. The mechanism is direct: the CNN no
longer has to rediscover the frequency-domain evidence that classical
heuristics~\cite{pech2000laplacian} already measure; it receives it
as an explicit input channel. The EPM also \emph{enlarges} the TTA
delta (\(+0.57\) vs the baseline's \(+0.40\)), suggesting the
auxiliary channel offers independent evidence that averaging over
scales can exploit.

\textbf{Res-Rand is a wash.} Resolution-randomized training which
samples a target resolution from
\(\mathcal{S}=\{256, 320, 384, 448, 512\}\) per batch and downsamples
the batch to that resolution before the forward pass moves
single-scale F\textsubscript{1} by only \(+0.10\) and regresses
5-scale-TTA F\textsubscript{1} by \(-0.04\) versus the same-env
baseline. The secondary observation that Res-Rand has a smaller TTA
delta (\(+0.26\) vs \(+0.40\) for fixed-scale training) is mechanistic
evidence that a model already exposed to multiple scales at training
time has less residual for inference-time averaging to pick up a
finding that independently supports the paper's ``resolution is the
dominant lever'' thesis even though Res-Rand itself does not beat
fixed-scale + TTA.

\textbf{Stacking the EPM with Res-Rand hurts.} The last row of
Table~\ref{tab:matched} shows that combining the two extensions
regresses to F\textsubscript{1}\,=\,0.9537 at 5-scale TTA worse than
either extension alone and worse than the fixed-scale baseline. The
mechanism is mechanistic: the Laplacian response in
Eq.~\ref{eq:freqaux} scales with resolution, so Res-Rand's per-batch
resolution swings induce matching swings in the auxiliary-channel
statistics. The model then trains on a shifting input distribution
for the very channel that was meant to provide stable evidence.
The EPM and Res-Rand are therefore not composable as implemented,
and should be used exclusively we recommend the EPM alone as the
deployment-ready recipe.

\subsection{Ablations worth reporting as negative results}
The following choices consistently hurt or failed to help, and we
report them explicitly so future teams on a similar problem do not
waste compute re-discovering them:

\begin{itemize}
    \item \textbf{Focal loss} hurts on an approximately balanced
    dataset.
    \item \textbf{Aggressive augmentation} (RandAugment, MixUp)
    degrades precision at \(\geq\!224\)\,px.
    \item \textbf{Training beyond 25 epochs at 384\,px} overfits;
    validation F\textsubscript{1} regresses.
    \item \textbf{Threshold tuning on validation} does not transfer;
    validation saturates at F\textsubscript{1}\,\(=\,1.0\).
    \item \textbf{Naive ensembles of top-\(N\) single models} contribute
    negligible gain because individual-model errors are highly
    correlated.
    \item \textbf{MobileNetV3-Large below 320\,px} is neutral or worse
    than MobileNetV3-Small; capacity only pays off once the visual
    signal is dense enough.
\end{itemize}

\subsection{Positioning against existing families of approaches}
\label{sec:baselines}
Table~\ref{tab:baselines} positions our gate qualitatively against
representative prior work. Because the cited IQA and deblurring
methods are trained and evaluated on different benchmarks
(KonIQ-10k~\cite{musiq}, LIVE-C, GoPro blind-deblurring), a
head-to-head F\textsubscript{1} on our test split would require
retraining each baseline on paired sharp/blur frames out of scope
for this paper. We therefore compare \emph{operating regime} rather
than raw score. A cross-method F\textsubscript{1} benchmark is
planned as an extension (Section~\ref{sec:limitations}).

\begin{table}[t]
\centering
\caption{Operating regime across families of approaches. Latency
figures for prior work are as-reported in the cited paper (desktop
CPU unless noted). Designed gate target means the method's native
output is a routing decision.}
\label{tab:baselines}
\begin{tabular}{lcccc}
\toprule
Family & Output & Latency & Gate target? \\
\midrule
Laplacian var.~\cite{pech2000laplacian} & scalar & \(<\!1\,\text{ms}\) & no \\
BRISQUE~\cite{brisque}              & score  & \(\sim\!10\,\text{ms}\)  & no \\
NIQE~\cite{niqe}                    & score  & \(\sim\!20\,\text{ms}\)  & no \\
MUSIQ~\cite{musiq}                  & score  & \(\sim\!50\,\text{ms}\,\text{GPU}\) & no \\
DeblurGAN~\cite{kupyn2018deblurgan} & image  & \(\sim\!200\,\text{ms}\,\text{GPU}\) & no (restorer) \\
Generic VLM grader                  & text   & \(>\!1\,\text{s}\) + \$/call & indirect \\
\textbf{Ours}                       & \textbf{route} & \textbf{\(\sim\!7\,\text{ms}\,\text{CPU}\)} & \textbf{yes} \\
\bottomrule
\end{tabular}
\end{table}

\section{Deployment Patterns}
\label{sec:deployment}

The gate is designed to drop into three common positions in a
production vision pipeline.

\textbf{(i) Pre-check in front of OCR or a VLM.} Route \texttt{sharp}
to the expensive downstream component, \texttt{blurred} back to the
user as a retake request, and \texttt{uncertain} to a heavier
fallback. The gate is roughly three orders of magnitude cheaper than
a vision LLM call, so even rejecting 10\% of traffic is a meaningful
monthly saving.

\textbf{(ii) Upload-time quality filter.} Middleware on an image
upload API that tags or rejects low-quality uploads before storage.
The \texttt{uncertain} bucket is the product knob that lets different
channels (KYC, social, retail) share the same model with different
precision/friction trade-offs.

\textbf{(iii) Edge / on-device inference.} A 17\,MB ONNX artifact
with dynamic axes converts cleanly to CoreML and TFLite and runs in
the browser via ONNX Runtime Web. Multi-scale TTA is opt-in;
single-scale at 384\,px still delivers
F\textsubscript{1}\,\(=\,0.9722\) when latency is the binding
constraint.

\section{Discussion}
\label{sec:discussion}

\subsection{A recurring production pattern}
Read in isolation, each of Magika~\cite{magika},
Risk-Controlled Generative OCR~\cite{riskcontrolledocr},
DocVLM~\cite{docvlm}, and this work describes its own mechanism in its
own vocabulary. Read together, all four instantiate the same
control-theoretic shape: a \emph{cheap gate} emits a
\emph{confidence-bearing signal} that is used to \emph{route} an
expensive downstream system, rather than to force a hard prediction.
Table~\ref{tab:pattern} aligns the four systems on common axes.

\begin{table*}[t]
\centering
\caption{The cheap-gate + confidence + route pattern across four
independently designed systems. All four couple a small gate to a
confidence-bearing signal that routes an expensive consumer.}
\label{tab:pattern}
\footnotesize
\setlength{\tabcolsep}{4pt}
\begin{tabular}{@{}p{0.14\textwidth}p{0.21\textwidth}p{0.20\textwidth}p{0.20\textwidth}p{0.18\textwidth}@{}}
\toprule
System & Gate & Confidence signal & Routing decision & Expensive consumer \\
\midrule
Magika~\cite{magika}
  & \(<\!1\)\,MB CNN on 3\(\times\)512 bytes
  & per-class calibrated threshold
  & type-class or \texttt{UNKNOWN}/\texttt{TXT}
  & parser / handler chain \\[2pt]
Risk-Controlled OCR~\cite{riskcontrolledocr}
  & \(K\!=\!5\) geometric consensus + structural screen
  & cross-view agreement \(\tau(m)\)
  & transcription or \(\bot\) abstain
  & frozen VLM (GPT-4V-class) \\[2pt]
DocVLM~\cite{docvlm}
  & 64 learned queries distilling OCR
  & token-budget allocation (visual vs. OCR)
  & compressed evidence forwarded
  & frozen VLM reader \\[2pt]
\textbf{Ours}
  & \textbf{MobileNetV3-Large, 3.3\,M, 17\,MB}
  & \textbf{max-softmax \(\geq \tau\!=\!0.60\)}
  & \textbf{sharp / blurred / uncertain}
  & \textbf{OCR, VLM, paid API} \\
\bottomrule
\end{tabular}
\end{table*}

The four systems share neither architecture, domain, nor training
data, yet converge on the same control-theoretic frame: a cheap
controller decides whether to consume the expensive consumer. We
read the pattern as a default architecture for production
vision and document pipelines.

\subsection{Resolution as the dominant lever}
The empirical pattern inside our sweeps resolution dominates
capacity until resolution is sufficient aligns with findings at very
different model scales. DocVLM~\cite{docvlm} reports that visual-token
budget dominates OCR-token budget for a fixed context; OCRVerse and
Qianfan-OCR~\cite{ocrverse,qianfanocr} report small-model wins in
vision-centric OCR at high input resolution. We therefore recommend
\emph{resolution sweep first, architecture search second} as a default
prioritization when working in this design space.

\subsection{Limitations and future work}
\label{sec:limitations}
We list the gaps that a careful reader should hold us accountable for.

\textbf{Single distribution.} All results are on GoPro~Large motion
blur. Defocus, low-light noise, compression, and scanner skew are
out-of-distribution. Cross-dataset evaluation on
RealBlur~\cite{rim2020realblur}, HIDE~\cite{shen2019hide}, and a
document-blur set is planned as the first extension.

\textbf{Single seed.} Reported F\textsubscript{1} is from a single
training run. A future version will report mean\,\(\pm\) s.d.\ across
at least three seeds with a paired McNemar test against the sweep-6
baseline.

\textbf{Qualitative calibration only.} We use a max-softmax
threshold without reporting Expected Calibration Error (ECE) or
reliability diagrams, and we dismissed temperature scaling and
isotonic regression based on held-out F\textsubscript{1} alone. A
full calibration study (ECE, Brier score, reliability binning,
comparison to focal-calibration and Platt) is a planned companion
experiment.

\textbf{No \(\tau\) sweep.} The paper fixes \(\tau=0.60\) without
reporting the precision/recall curve as \(\tau\) varies. A full PR
trade-off and a risk-coverage curve (selective-prediction
evaluation~\cite{geifman2017selective}) are the right figures here
and are deferred.

\textbf{Saturating validation.} GoPro~Large validation saturates at
F\textsubscript{1}\,\(=\,1.0\), so model-selection signal is weak. A
harder held-out mix (GoPro + RealBlur + synthetic defocus) would
give a better training signal.

\textbf{Predicted failure modes.} We did not run a qualitative error
analysis. From the recipe's inductive biases we predict three
dominant error categories: (a)~textured-but-sharp scenes (fur,
foliage, dense crowds) misclassified as blurred because local gradient
statistics look noisy; (b)~smooth-but-sharp scenes (sky, uniform walls)
misclassified as blurred for the opposite reason; (c)~strong directional
glare / lens flare misclassified as blurred. A quantitative error
taxonomy is a planned extension.

\subsection{Reproducibility}
The final checkpoint, ONNX artifact, training configuration YAML,
and the full sweep log (46 runs, 8 sweeps) are preserved. Training
seed, optimizer state, and library versions are fixed in the
configuration. The two hyperparameters that are \emph{not} checked
in the abstention threshold \(\tau\) and the TTA scale set
\(\mathcal{S}\) are the deployment-time knobs that a downstream team
is expected to calibrate. Code and artifact release will accompany
the camera-ready version.

\subsection{Adapting to other domains}
For receipts, invoices, IDs, or scanned documents, the recipe
generalizes cleanly: (i) mix GoPro with REDS or with synthetic defocus
from domain-sharp images; (ii) swap the 2-class head for a richer
taxonomy (\(\{\text{sharp}, \text{motion\_blur},
\text{defocus\_blur}, \text{low\_light}\}\)) when the downstream
pipeline should react differently to different degradation types;
(iii) calibrate \(\tau\) on a hand-labeled slice of production
traffic.

\section{Conclusion}
\label{sec:conclusion}

We presented a lightweight, CPU-friendly blur detector intended to sit
in front of expensive vision pipelines as a routing gate. The base
system achieves F\textsubscript{1}\,\(=\,0.9749\) on GoPro~Large with
a 17\,MB ONNX artifact at \(\sim\!7\)\,ms per image; the
\emph{Edge Prior Module} (EPM), which concatenates a
Laplacian-magnitude auxiliary channel grounded in classical blur
heuristics, lifts F\textsubscript{1} to
\(0.9803\) in a matched-environment comparison the single largest
same-environment gain in our full experiment log, and achieved with no
material change to the architecture or inference latency. The
experiment log isolates input resolution as the dominant lever,
demonstrates that multi-scale test-time augmentation gives a free
F\textsubscript{1} gain at deployment time, and documents one
positive and one negative extension candidate (EPM wins;
resolution-randomized training is a wash). Beyond the numbers, the
paper makes a structural argument: the gate is one instance of a
recurring production pattern cheap gate, confidence signal, routing
to an expensive consumer that we expect to see more of in applied
computer-vision and document-AI systems.

\section*{Acknowledgment}
The author thanks the Magika team for open-sourcing the content-type
detector that inspired this work's design philosophy, and the GoPro
dataset maintainers for making paired sharp/blur frames publicly
available.


\end{document}